\documentclass[11pt]{article}

\usepackage{acl}
\usepackage{times}
\usepackage{latexsym}
\usepackage[T1]{fontenc}
\usepackage[utf8]{inputenc}
\usepackage{inconsolata}
\usepackage{graphicx}
\usepackage{booktabs}
\usepackage{multirow}
\usepackage{array}
\usepackage{makecell}
\usepackage{tikz}
\usetikzlibrary{positioning,arrows.meta,calc,backgrounds,shapes.geometric}
\usepackage{xcolor}
\usepackage[most]{tcolorbox}
\tcbuselibrary{breakable,skins}
\usepackage{enumitem}
\usepackage{amsmath}
\usepackage{amssymb}

\definecolor{tabhead}{HTML}{E8E8E8}
\definecolor{best}{HTML}{0e9692}    
\definecolor{ours}{HTML}{16d9d2}    
\definecolor{warn}{HTML}{B22222}
\definecolor{judge}{HTML}{6A4C9C}   
\definecolor{cpath}{HTML}{C97A2C}   
\definecolor{mrgnode}{HTML}{0ecdb4}  
\definecolor{promptA}{HTML}{82c4ba}  
\definecolor{promptB}{HTML}{d08770}  
\definecolor{promptC}{HTML}{afc29d}  
\definecolor{promptD}{HTML}{ebcb8b}  

\newcommand{\best}[1]{\textbf{#1}}
\newcommand{\sub}[1]{\textsubscript{#1}}
\newcommand{\warntext}[1]{\textcolor{warn}{#1}}

\newcommand{\dpos}[1]{\textcolor{mrgnode!75!black}{\scriptsize $#1$}}
\newcommand{\dneg}[1]{\textcolor{warn}{\scriptsize $#1$}}

\title{Dependency-Aware Trajectory Refinement for Efficient Multi-Turn Agent Fine-Tuning}

\author{
  Zhuo Chen$^{1}$, Zhen Zhang, Xinyu Wang, Kewei Tu$^{1}$\thanks{Corresponding author} \\
  $^{1}$School of Information Science and Technology, ShanghaiTech University \\
  $^{1}$Shanghai Engineering Research Center of Intelligent Vision and Imaging \\
  \texttt{chenzhuo@shanghaitech.edu.cn}
}

\begin{document}
\maketitle

\begin{abstract}
Multi-turn agent trajectories often
contain redundant rounds (failed tool calls, parallel sub-queries,
verification-only steps) that inflate both training and inference
cost. We propose viewing each trajectory as a \emph{round-level
dependency DAG} that exposes which rounds are globally load-bearing
for the final answer, and fine-tune agents on trajectories refined
through this DAG. Given an LLM-annotated DAG, these edits are deterministic and interpretable, with optional rephrasing.
Models trained on these refined trajectories consistently outperform those trained on the original trajectories at lower inference cost.
Specifically, across four multi-modal QA
benchmarks, our refinements improve downstream accuracy by up to
$1.7$\,pp over vanilla SFT (and $5.7$\,pp over an LLM-deletion
baseline) while reducing per-sample inference messages by up to
approximately $40\%$ and inference tokens by up to approximately $48\%$, translating
to substantial savings in compute and serving cost. Code is available \href{https://github.com/Chord-Chen-30/LLaMA-Factory-lite}{here}.
\end{abstract}

\section{Introduction}
\label{sec:intro}

Tool-augmented multi-modal agents that interleave reasoning with
retrieval calls now define the state of the art for visual question
answering~\citep{yao2023react,schick2023toolformerlanguagemodelsteach,qin2023toolllmfacilitatinglargelanguage,deng2023mind2web,OSWorld}.
Training such agents from a pretrained model typically needs high-quality multi-turn trajectories that demonstrate how to use tools to solve complex problems. These trajectories are valuable but not always information-dense. While synthesizing these trajectories, the source model inevitably runs failed searches, double-checks already confirmed facts, or branches into parallel sub-queries whose results never enter the final answer. Cost then compounds at both training and inference stages.

A natural temptation is to compress trajectories before SFT. We focus on two contrasting operations. The first intuitive baseline method is \textbf{round deletion} without the contextual awareness of a dependency DAG. An LLM judge marks middle rounds as redundant towards the final answer. Through experiments, we find it aggressive but lossy, since removing a round whose facts are still cited leaves the student ``hallucinating sources''. In contrast, our proposed method operates on a dependency graph. We first construct a DAG over rounds, then perform two levels of structural pruning. The first level removes non-terminal leaf nodes, and the second level merges independent sibling rounds.

\begin{figure}[tb]
\centering
\includegraphics[width=0.97\linewidth]{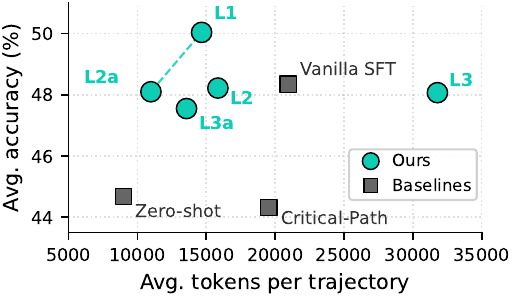}
\caption{\textbf{Accuracy vs.\ inference-cost trade-off.} \textbf{L1} reaches the highest accuracy. \textbf{L2a} halves the inference tokens at vanilla-SFT accuracy.}
\label{fig:pareto}
\end{figure}

\definecolor{chainbg}{HTML}{EEF1F5}
\definecolor{depbg}{HTML}{E1F2EE}
\definecolor{afterbg}{HTML}{EBF2DE}
\definecolor{depgreen}{HTML}{0ecdb4}  
\newcommand{\seqarrow}{\protect\tikz[baseline=-0.4ex]\protect\draw[->,>={Stealth[length=3pt]},black,dash pattern=on 1.5pt off 1pt,line width=0.6pt] (0,0)--(0.55,0);}
\newcommand{\deparrow}{\protect\tikz[baseline=-0.4ex]\protect\draw[->,>={Stealth[length=4.5pt]},color=depgreen,line width=1.2pt] (0,0)--(0.55,0);}
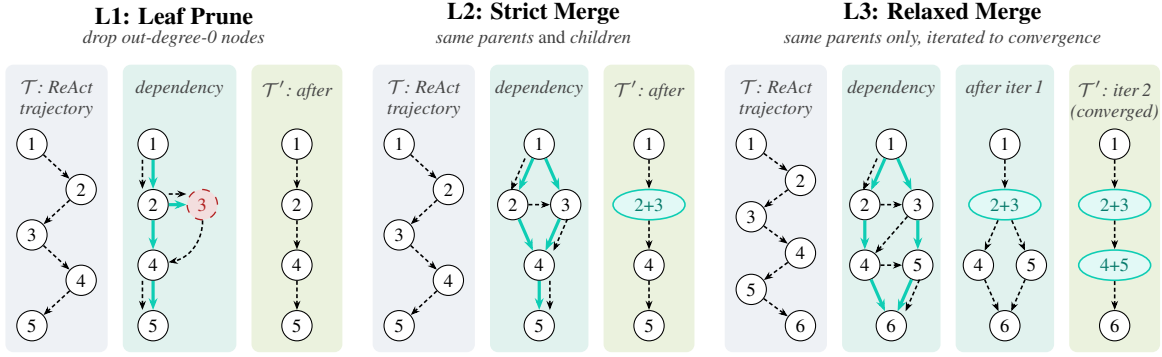
\begin{figure*}[!t]
\centering
\begin{tikzpicture}[
  every node/.style={font=\scriptsize},
  rd/.style={circle,draw=black,fill=white,minimum size=4mm,inner sep=0pt},
  drop/.style={circle,draw=warn,fill=warn!15,minimum size=4mm,inner sep=0pt,text=warn,dashed,line width=0.5pt},
  mrg/.style={ellipse,draw=mrgnode,fill=mrgnode!12,minimum width=9.5mm,minimum height=4mm,inner sep=0pt,text=mrgnode!60!black,line width=0.6pt},
  seq/.style={->,>={Stealth[length=3.5pt]},black,dash pattern=on 1.5pt off 1pt,line width=0.6pt},
  dep/.style={->,>={Stealth[length=4.5pt]},color=depgreen,line width=1.2pt},
  panel/.style={font=\footnotesize\bfseries},
  crit/.style={font=\scriptsize\itshape,text=gray!50!black},
  sublbl/.style={font=\scriptsize\itshape,text=gray!55!black}
]

\begin{scope}[xshift=0cm]
  \node[panel,anchor=south] at (2.225, 2.45) {\textbf{L1: Leaf Prune}};
  \node[crit,anchor=south]  at (2.225, 2.15) {drop out-degree-0 nodes};

  \path[fill=chainbg,rounded corners=3pt] (0.00,-1.75) rectangle (1.35, 2.05);
  \node[sublbl,anchor=north,align=center] at (0.675, 1.98) {$\mathcal{T}$:\ ReAct\\trajectory};
  \node[rd] (alTR1) at (0.35, 1.00) {1};
  \node[rd] (alTR2) at (1.00, 0.40) {2};
  \node[rd] (alTR3) at (0.35,-0.20) {3};
  \node[rd] (alTR4) at (1.00,-0.80) {4};
  \node[rd] (alTR5) at (0.35,-1.40) {5};
  \draw[seq] (alTR1)--(alTR2);
  \draw[seq] (alTR2)--(alTR3);
  \draw[seq] (alTR3)--(alTR4);
  \draw[seq] (alTR4)--(alTR5);

  \path[fill=depbg,rounded corners=3pt] (1.55,-1.75) rectangle (3.05, 2.05);
  \node[sublbl,anchor=north] at (2.30, 1.98) {dependency};
  \node[rd]   (alR1) at (1.95, 1.00) {1};
  \node[rd]   (alR2) at (1.95, 0.20) {2};
  \node[drop] (alR3) at (2.60, 0.20) {3};
  \node[rd]   (alR4) at (1.95,-0.60) {4};
  \node[rd]   (alR5) at (1.95,-1.40) {5};
  \draw[dep] (alR1)--(alR2);
  \draw[dep] (alR2)--(alR3);
  \draw[dep] (alR2)--(alR4);
  \draw[dep] (alR4)--(alR5);
  \draw[seq,transform canvas={xshift=-4pt}] (alR1)--(alR2);
  \draw[seq,transform canvas={yshift=4pt}]  (alR2)--(alR3);
  \draw[seq] (alR3) to[bend left=40] (alR4);
  \draw[seq,transform canvas={xshift=-4pt}] (alR4)--(alR5);

  \path[fill=afterbg,rounded corners=3pt] (3.25,-1.75) rectangle (4.45, 2.05);
  \node[sublbl,anchor=north,align=center] at (3.85, 1.98) {$\mathcal{T}'$:\ after};
  \node[rd] (arR1) at (3.85, 1.00) {1};
  \node[rd] (arR2) at (3.85, 0.20) {2};
  \node[rd] (arR4) at (3.85,-0.60) {4};
  \node[rd] (arR5) at (3.85,-1.40) {5};
  \draw[seq] (arR1)--(arR2);
  \draw[seq] (arR2)--(arR4);
  \draw[seq] (arR4)--(arR5);
\end{scope}

\begin{scope}[xshift=4.85cm]
  \node[panel,anchor=south] at (2.125, 2.45) {\textbf{L2: Strict Merge}};
  \node[crit,anchor=south]  at (2.125, 2.15) {same parents \emph{and} children};

  \path[fill=chainbg,rounded corners=3pt] (0.00,-1.75) rectangle (1.35, 2.05);
  \node[sublbl,anchor=north,align=center] at (0.675, 1.98) {$\mathcal{T}$:\ ReAct\\trajectory};
  \node[rd] (blTR1) at (0.35, 1.00) {1};
  \node[rd] (blTR2) at (1.00, 0.40) {2};
  \node[rd] (blTR3) at (0.35,-0.20) {3};
  \node[rd] (blTR4) at (1.00,-0.80) {4};
  \node[rd] (blTR5) at (0.35,-1.40) {5};
  \draw[seq] (blTR1)--(blTR2);
  \draw[seq] (blTR2)--(blTR3);
  \draw[seq] (blTR3)--(blTR4);
  \draw[seq] (blTR4)--(blTR5);

  \path[fill=depbg,rounded corners=3pt] (1.55,-1.75) rectangle (2.85, 2.05);
  \node[sublbl,anchor=north] at (2.20, 1.98) {dependency};
  \node[rd] (blR1) at (2.20, 1.00) {1};
  \node[rd] (blR2) at (1.85, 0.20) {2};
  \node[rd] (blR3) at (2.55, 0.20) {3};
  \node[rd] (blR4) at (2.20,-0.60) {4};
  \node[rd] (blR5) at (2.20,-1.40) {5};
  \draw[dep] (blR1)--(blR2);
  \draw[dep] (blR1)--(blR3);
  \draw[dep] (blR2)--(blR4);
  \draw[dep] (blR3)--(blR4);
  \draw[dep] (blR4)--(blR5);
  \draw[seq,transform canvas={xshift=-3pt}] (blR1)--(blR2);
  \draw[seq] (blR2)--(blR3);
  \draw[seq,transform canvas={xshift=3pt}]  (blR3)--(blR4);
  \draw[seq,transform canvas={xshift=4pt}]  (blR4)--(blR5);

  \path[fill=afterbg,rounded corners=3pt] (3.05,-1.75) rectangle (4.25, 2.05);
  \node[sublbl,anchor=north,align=center] at (3.65, 1.98) {$\mathcal{T}'$:\ after};
  \node[rd]  (brR1)  at (3.65, 1.00) {1};
  \node[mrg] (brR23) at (3.65, 0.20) {2+3};
  \node[rd]  (brR4)  at (3.65,-0.60) {4};
  \node[rd]  (brR5)  at (3.65,-1.40) {5};
  \draw[seq] (brR1)--(brR23);
  \draw[seq] (brR23)--(brR4);
  \draw[seq] (brR4)--(brR5);
\end{scope}

\begin{scope}[xshift=9.50cm]
  \node[panel,anchor=south] at (2.875, 2.45) {\textbf{L3: Relaxed Merge}};
  \node[crit,anchor=south]  at (2.875, 2.15) {same parents only, iterated to convergence};

  \path[fill=chainbg,rounded corners=3pt] (0.00,-1.75) rectangle (1.35, 2.05);
  \node[sublbl,anchor=north,align=center] at (0.675, 1.98) {$\mathcal{T}$:\ ReAct\\trajectory};
  \node[rd] (clTR1) at (0.35, 1.00) {1};
  \node[rd] (clTR2) at (1.00, 0.52) {2};
  \node[rd] (clTR3) at (0.35, 0.04) {3};
  \node[rd] (clTR4) at (1.00,-0.44) {4};
  \node[rd] (clTR5) at (0.35,-0.92) {5};
  \node[rd] (clTR6) at (1.00,-1.40) {6};
  \draw[seq] (clTR1)--(clTR2);
  \draw[seq] (clTR2)--(clTR3);
  \draw[seq] (clTR3)--(clTR4);
  \draw[seq] (clTR4)--(clTR5);
  \draw[seq] (clTR5)--(clTR6);

  \path[fill=depbg,rounded corners=3pt] (1.55,-1.75) rectangle (2.85, 2.05);
  \node[sublbl,anchor=north] at (2.20, 1.98) {dependency};
  \node[rd] (clR1) at (2.20, 1.00) {1};
  \node[rd] (clR2) at (1.85, 0.20) {2};
  \node[rd] (clR3) at (2.55, 0.20) {3};
  \node[rd] (clR4) at (1.85,-0.60) {4};
  \node[rd] (clR5) at (2.55,-0.60) {5};
  \node[rd] (clR6) at (2.20,-1.40) {6};
  \draw[dep] (clR1)--(clR2);
  \draw[dep] (clR1)--(clR3);
  \draw[dep] (clR2)--(clR4);
  \draw[dep] (clR3)--(clR5);
  \draw[dep] (clR4)--(clR6);
  \draw[dep] (clR5)--(clR6);
  \draw[seq,transform canvas={xshift=-3pt}] (clR1)--(clR2);
  \draw[seq] (clR2)--(clR3);
  \draw[seq] (clR3)--(clR4);
  \draw[seq] (clR4)--(clR5);
  \draw[seq,transform canvas={xshift=3pt}]  (clR5)--(clR6);

  \path[fill=depbg,rounded corners=3pt] (3.05,-1.75) rectangle (4.35, 2.05);
  \node[sublbl,anchor=north] at (3.70, 1.98) {after iter\,1};
  \node[rd]  (cmR1)  at (3.70, 1.00) {1};
  \node[mrg] (cmR23) at (3.70, 0.20) {2+3};
  \node[rd]  (cmR4)  at (3.35,-0.60) {4};
  \node[rd]  (cmR5)  at (4.05,-0.60) {5};
  \node[rd]  (cmR6)  at (3.70,-1.40) {6};
  \draw[seq] (cmR1)--(cmR23);
  \draw[seq] (cmR23)--(cmR4);
  \draw[seq] (cmR23)--(cmR5);
  \draw[seq] (cmR4)--(cmR6);
  \draw[seq] (cmR5)--(cmR6);

  \path[fill=afterbg,rounded corners=3pt] (4.55,-1.75) rectangle (5.75, 2.05);
  \node[sublbl,anchor=north,align=center] at (5.15, 1.98) {$\mathcal{T}'$:\ iter\,2\\(converged)};
  \node[rd]  (crR1)  at (5.15, 1.00) {1};
  \node[mrg] (crR23) at (5.15, 0.20) {2+3};
  \node[mrg] (crR45) at (5.15,-0.60) {4+5};
  \node[rd]  (crR6)  at (5.15,-1.40) {6};
  \draw[seq] (crR1)--(crR23);
  \draw[seq] (crR23)--(crR45);
  \draw[seq] (crR45)--(crR6);
\end{scope}

\end{tikzpicture}
\caption{Three levels of refinements on a structural dependency DAG. For each level, \emph{trajectory} traces the recorded round order (\seqarrow), \emph{dependency} overlays the actual data dependencies on the same nodes (\deparrow), and \emph{after} shows the rewritten trajectory. Round~1 is the user query, the highest-indexed round the assistant answer.}
\label{fig:dag}
\end{figure*}

Once a trajectory is recast as a round-level dependency DAG, edits
become \emph{deterministic, interpretable, and inexpensive to
apply}. We instantiate three such variants of increasing aggressiveness
(leaf prune, strict merge, relaxed merge), each preserving the
load-bearing dependencies of the original trajectory, optionally
followed by LLM rephrasing. Our results reveal two complementary
sweet spots (Fig.~\ref{fig:pareto}). Leaf prune (\textbf{L1}) anchors the
high-accuracy end at $+5.7$\,pp over the Critical-Path baseline.
Strict merge with LLM rephrasing (\textbf{L2a}) anchors the cost-efficient
end, matching vanilla-SFT accuracy while using only $11.0$\,K
inference tokens per sample, vs.\ $20.9$\,K for vanilla SFT.
Together these two points define the best accuracy--cost trade-off
among the SFT systems we compare. We further note that the most
aggressive variant, relaxed merge (\textbf{L3}), is sensitive to the
one-tool-call-per-turn protocol used at inference. A sizeable
fraction of samples exhibit extended interaction loops at noticeably
higher inference cost. Structural edits work best when they preserve the
train--test interaction pattern expected at inference.

\noindent\textbf{Contributions.}
(1)~A dependency-DAG view of agent trajectories that turns trajectory
refinement into a transparent, deterministic graph-editing problem,
with three concrete levels of structural edit.
(2)~Across four multi-modal QA benchmarks, our refined trajectories train models that surpass vanilla-SFT accuracy at substantially lower inference cost.
(3)~An LLM-rephrasing extension that closes the training/inference format gap from round merging, together with an analysis linking each edit level's behaviour to that gap.

\section{Method}
\label{sec:method}

\subsection{Preliminary}
\label{sec:notation}
We discuss agent trajectories in the ReAct paradigm \citep{yao2023react}, where an agent interleaves reasoning and acting to solve tasks. 
An agent trajectory $\mathcal{T}$ is a sequence of \emph{rounds}.
Round\,1 is the user's question. Rounds\,$2..N{-}1$ each consist of
an assistant turn (\texttt{<think>}+\texttt{<tool\_call>}) followed by a
user turn (\texttt{<tool\_response>}). Round\,$N$ is the final assistant
answer (\texttt{<think>}+\texttt{<answer>}). We refine $\mathcal{T}$ into a $\mathcal{T}'$ that preserves the final answer while containing no more rounds than $\mathcal{T}$.

\subsection{Round-Level Dependency DAG}
\label{sec:dag}
For each trajectory $\mathcal{T}$ we extract a DAG $G(\mathcal{T})$
whose edges $i \to j$ record \emph{globally load-bearing}
dependencies. Round~$i$ produced a concrete artefact (number, named
entity, URL, intermediate conclusion) that Round~$j$ uses to reach
the final answer. Edges encoding mere narrative
reference (``... was unhelpful, let me try ...'')
are excluded. After parsing, any cycles are removed by depth-first
traversal. We obtain these edges by querying GPT-5.4 with
the prompt shown in Sec.~\ref{app:prompts}.

\subsection{Three Levels of Structural Edit}
\label{sec:levels}

We apply three edits of increasing aggressiveness, illustrated in
Fig.~\ref{fig:dag}:

\paragraph{Leaf Prune (L1).} Iteratively remove every out-degree-zero
round except round~1 (the user query) and round~$N$ (the assistant
answer). These nodes correspond to dead-end tool calls (failed
retrievals, abandoned sub-queries) whose outputs feed nothing
downstream.

\paragraph{Strict Merge (L2).} Two rounds with the same upstream
context and the same downstream consumer are interchangeable in
dataflow. From the perspective of the final answer, they are
parallel computations of the same logical step. \textbf{L2} fuses these
sibling rounds into one. This collapses redundant parallelism while
preserving every dependency in the DAG. \textbf{L2} is applied on top of \textbf{L1}.

\paragraph{Relaxed Merge (L3).} \textbf{L3} relaxes \textbf{L2}'s criterion by
requiring only shared parents, regardless of downstream consumers,
and applies the rule iteratively. Because every merge changes the
parent set of nodes downstream of it, new sibling pairs can emerge
after a pass, so we keep fusing until no candidates remain. In
Fig.~\ref{fig:dag} (panel \textbf{L3}), merging $\{2,3\}$ into $2{+}3$
makes rounds $4$ and $5$ share the new parent $2{+}3$, and they are
fused into $4{+}5$ in a second pass. \textbf{L3} is applied on top of \textbf{L1}.

\paragraph{Canonical Format.} All edits keep one outer tag per turn
(\texttt{<think>} + \texttt{<tool\_call>}/\texttt{<answer>} on the
assistant, \texttt{<tool\_response>} on the user). Merged bodies are
concatenated inside the tag.

\subsection{LLM-Rephrased Merge (L2a, L3a)}
\label{sec:rephrase}
A merged round's \texttt{<think>} is a concatenation of several
disjoint trains of thought, a pattern absent in original
trajectories. \textbf{L2a} and \textbf{L3a} apply an LLM rewrite to the merged
\texttt{<think>} of \textbf{L2} and \textbf{L3} that produces a single coherent passage
while preserving every sub-task transition and tool-call reference
(prompt in Sec.~\ref{app:prompts}). The \texttt{<tool\_call>} and
\texttt{<tool\_response>} bodies are kept verbatim.

\section{Experiments}
\label{sec:experiments}

\subsection{Setup}
\label{sec:setup}

\paragraph{Training Data.} We use 6196 multi-modal, tool-using
trajectories extended from
\citet{geng2025webwatcherbreakingnewfrontier}. Table~\ref{tab:trainstats} reports message and token reductions for
each edit on a \emph{fair-comparison subset}, 1334
trajectories that at least one edit modifies.
\textbf{Critical-Path} is the most aggressive, and \textbf{L1} is the most conservative. \textbf{L2} and \textbf{L3} fall in between where \textbf{L3} removes more rounds but fewer tokens than \textbf{L2}.

\paragraph{Systems Compared.}
(i) \textbf{Zero-shot}: the base \texttt{Qwen3-VL-30B-A3B-Thinking}
model without SFT~\citep{bai2025qwen3vltechnicalreport}.
(ii) \textbf{Vanilla SFT}: SFT on the unmodified 6196 raw trajectories.
(iii) \textbf{Critical-Path}: an LLM marks each middle round
to be ``Keep or Remove'', with a grounding check that vetoes deletions that would
orphan named entities in the final answer.
(iv) \textbf{Ours}: SFT on the same base with \textbf{L1}–\textbf{L3a} trajectory refinement. See Secs.~\ref{app:train} and \ref{app:inference} for full settings.

\paragraph{Benchmarks.}We evaluate on four multi-modal, tool-using QA datasets. SimpleVQA (visual factuality, 300; \citealp{cheng2025simplevqamultimodalfactualityevaluation}), LiveVQA (recent-knowledge visual QA, 300; \citealp{fu2025seekingupdatinglivevisual}), HLE (hard knowledge problems, 330; \citealp{phan2026humanitysexam}), and MMSearch (multi-modal search, 171; \citealp{jiang2024mmsearch}). We report \texttt{gpt-5-nano} judge accuracy along with two efficiency proxies.

\begin{table}[tb]
\centering
\small
\setlength{\tabcolsep}{3pt}
\renewcommand{\arraystretch}{1.05}
\begin{tabular}{lcccc}
\toprule
Config & \makecell{Avg.\\\#msgs} & \makecell{Msg.\\red.}
       & \makecell{Avg.\\\#toks} & \makecell{Tok.\\red.} \\
\midrule
\emph{Original}        & 11.77 & ---     & 3{,}692 & ---     \\
Critical-Path    &  8.29 & 29.46\% & 2{,}781 & 24.65\% \\
\midrule
\textbf{L1}            &  9.63 & 18.18\% & 3{,}096 & 16.15\% \\
\textbf{L2}            &  9.22 & 21.61\% & 3{,}091 & 16.28\% \\
\textbf{L3}            &  8.87 & 24.67\% & 3{,}087 & 16.39\% \\
\textbf{L2a}           &  9.22 & 21.61\% & 3{,}080 & 16.57\% \\
\textbf{L3a}           &  8.87 & 24.67\% & 3{,}067 & 16.93\% \\
\bottomrule
\end{tabular}
\caption{Training-set message and token reduction.}
\label{tab:trainstats}
\end{table}

\subsection{Main Results}
\label{sec:main_results}

Table~\ref{tab:main} reports task accuracy, average message and token counts, and derived cost-effectiveness ratios across the four benchmarks. Overall, Leaf Prune (\textbf{L1}) achieves the highest accuracy, while the rephrased strict merge (\textbf{L2a}) emerges as the most cost-effective variant. Three key observations follow from these results:

\noindent(i) Breaking limits on both accuracy and efficiency. Our methods successfully decouple the tight coupling between high accuracy and high cost. \textbf{L1} establishes a new performance ceiling ($50.04\%$ Avg.\ Acc.), outperforming vanilla SFT by $+1.7$\,pp. Simultaneously, \textbf{L2a} redefines inference efficiency, achieving the highest accuracy-per-token ratio among all SFT variants while maintaining accuracy comparable to the baseline. Breaking the 4-benchmark average down, \textbf{L2a} is uniformly the cheapest in messages, and the cheapest in tokens, on every single benchmark among SFT systems (Sec.~\ref{app:detailed_perf}).

\begin{table*}[!t]
\centering
\footnotesize
\setlength{\tabcolsep}{3.5pt}
\renewcommand{\arraystretch}{1.0}
\begin{tabular}{@{}clcccccccc@{}}
\toprule
 & & \makecell{Zero\\-shot} & \makecell{Vanilla\\SFT}
 & \makecell{Critical\\Path}
 & \textbf{L1} & \textbf{L2} & \textbf{L3} & \textbf{L2a} & \textbf{L3a} \\
\midrule
\multirow{5}{*}{\emph{\makecell[c]{Task\\accuracy}}}
  & SimpleVQA  & 66.67 & 68.67 & 60.67 & \best{70.67} & 68.67 & 66.67 & 67.33 & 64.33 \\
  & LiveVQA    & 48.00 & 51.00 & 46.33 & \best{53.67} & 51.33 & 50.33 & 51.00 & 51.00 \\
  & HLE        &  8.48 &  9.39 & 11.21 & 10.30 & 10.30 & \best{11.52} & \best{11.52} &  8.79 \\
  & MMSearch   & 55.56 & 64.33 & 59.06 & 65.50 & 62.57 & 63.74 & 62.57 & \best{66.08} \\
  & \textbf{Avg.\ Acc.} & 44.68 & 48.35 & 44.32 & \best{50.04} & 48.22 & 48.07 & 48.10 & 47.55 \\
\midrule
\multirow{2}{*}{\emph{\makecell[c]{Inference\\efficiency}}}
  & Avg.\ \#msgs  & 41.67 & 23.94 & 27.33 & 15.90 & 22.25 & 30.19 & \best{14.30} & 16.63 \\
  & Avg.\ \#toks  & \best{8{,}977}\sub{\dag} & 20{,}940 & 19{,}540 & 14{,}657 & 15{,}841 & 31{,}779 & 10{,}977 & 13{,}766 \\
\midrule
\multirow{2}{*}{\emph{\makecell[c]{Cost-\\effectiveness}}}
  & Acc.\,/\,msg & 1.07 & 2.02 & 1.62 & 3.15 & 2.17 & 1.59 & \best{3.36} & 2.86 \\
  & Acc.\,/\,tok ($\times 10^{3}$) & 4.98\sub{\dag} & 2.31 & 2.27 & 3.41 & 3.04 & 1.51 & \best{4.38} & 3.45 \\
\bottomrule
\end{tabular}
\caption{Main results across four multi-modal QA benchmarks.
\best{Bold} marks the best in each row.
\sub{\dag}\,Artefactual: Zero-shot frequently enters stuck loops on
HLE (see Sec.~\ref{app:stuck}), deflating its token average.}
\label{tab:main}
\end{table*}

\noindent(ii) Flexible accuracy/cost selection under diverse deployment constraints.
When compute budgets are generous, \textbf{L1} serves as the optimal choice, leading performance on three of the four benchmarks. Conversely, under strict cost or latency constraints, \textbf{L2a} provides an ideal drop-in alternative, slashing token overhead by approximately 48\% without sacrificing accuracy.

\noindent(iii) Boundary exploration and specialized strengths. Relaxed merging (\textbf{L3}) increases token costs on standard tasks but achieves a peak accuracy of $11.52\%$ on the challenging HLE dataset, showing promise for complex reasoning. In contrast, the \textbf{Critical-Path} baseline drops sharply, confirming that bluntly deleting intermediate rounds hurts generalization.

\section{Analysis}
\label{sec:analysis}

In this section, first we analyse why relaxed merge raises stuck rates and how LLM rephrasing mitigates that gap. Second, we compare against Chain-of-Draft as an inference-time baseline. Third, we test whether L1's kept-node set is stable under alternative annotators.

\begin{figure}[tb]
\centering
\includegraphics[width=0.99\linewidth]{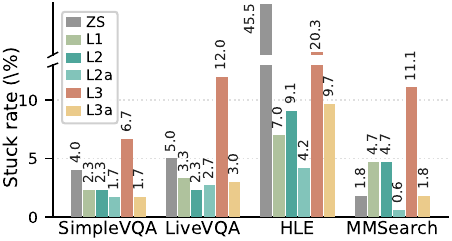}
\caption{\emph{Stuck rate} (\%), the fraction of samples that hit the 128-round agent-loop ceiling.}
\label{fig:stuck}
\end{figure}

\subsection{Trajectory Merging and Rephrasing}
When a merge group contains sibling nodes with divergent children, the training turn aggregates multiple tool responses into one \texttt{<tool\_response>} block. This format departs from typical multi-turn interactions and creates a shift that can disrupt the model's behavior during inference. As a result, samples hit the 128-round agent-loop ceiling $2$--$3\times$ as often under \textbf{L3} as under \textbf{L2} across all four benchmarks (Fig.~\ref{fig:stuck}). Excluding these stuck samples nearly aligns \textbf{L3}'s median message count with \textbf{L2}'s. \textbf{L2} avoids this distribution shift by restricting merges to interchangeable siblings.

\textbf{L2a} and \textbf{L3a} rewrite the concatenated \texttt{<think>} bodies into a single coherent reasoning passage. On \textbf{L3}, this rewrite both largely closes the stuck-rate gap (Fig.~\ref{fig:stuck}) and substantially cuts per-sample cost ($45\%$ messages, $57\%$ tokens) at comparable accuracy. The residual gap between \textbf{L3a} and \textbf{L2a} is structural, since \textbf{L3} fuses siblings whose downstream consumers differ, a pattern that rephrasing alone cannot reshape.

\begin{figure}[!t]
  \centering
  \includegraphics[width=0.99\linewidth]{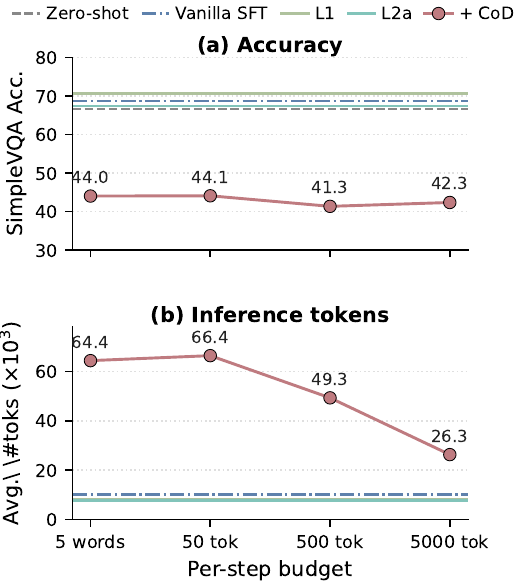}
  \caption{Chain-of-Draft sweep on SimpleVQA versus Zero-shot, Vanilla SFT, \textbf{L1}, and \textbf{L2a}.}
  \label{fig:cod}
  \end{figure}

\subsection{Comparison with Chain-of-Draft}
We also test Chain of Draft~\citep{xu2025chainofdraft}, a prompting-only method that compresses single-response chain of thought, as an inference-time baseline on the unmodified Zero-shot model with same agent loop and judge. Sweeping the per-step budget from $5$~words to $5000$~tokens on SimpleVQA (Fig.~\ref{fig:cod}), accuracy falls about $22$\,pp below Zero-shot while tokens rise to $6$--$8\times$ the Zero-shot average. CoD targets short single-turn reasoning; tightening per-round \texttt{<think>} in a multi-turn tool loop makes the agent fire tools before framing the sub-problem, so it becomes more wasteful. We therefore omit CoD from Table~\ref{tab:main}. Prompting-only compression for single-response tasks does not transfer here, which supports editing trajectories before training rather than constraining generation at inference.

\subsection{Annotator Agreement}
We re-annotate training trajectories with two alternative annotators and compare the kept-node set that L1 consumes against the GPT-5.4 reference (Table~\ref{tab:annotator}). Kept-P is the fraction of an alternative's kept rounds that GPT-5.4 also keeps. Kept-R is the fraction of GPT-5.4's kept rounds that the alternative retains. Both recover the same load-bearing core and differ mainly in how aggressively they prune.

\begin{table}[t]
\centering
\small
\begin{tabular}{lccc}
\toprule
Annotator & Kept-P & Kept-R & Kept-F1 \\
\midrule
\texttt{claude-sonnet-4-6} & 0.916 & 0.819 & 0.865 \\
\texttt{deepseek-v4-pro}   & 0.990 & 0.575 & 0.727 \\
\bottomrule
\end{tabular}
\caption{Kept-node agreement vs.\ GPT-5.4.}
\label{tab:annotator}
\end{table}

\subsection{Additional Experiments}
Appendix material covers training and inference settings (Secs.~\ref{app:train}--\ref{app:inference}), per-benchmark cost breakdowns (Sec.~\ref{app:detailed_perf}), stuck-rate notes (Sec.~\ref{app:stuck}), merge-format alignment (Sec.~\ref{app:merge_format}), DPO results and setting (Secs.~\ref{app:dpo}--\ref{app:dpo_setting}), checkpoint selection (Sec.~\ref{app:loss}), a qualitative DAG example (Sec.~\ref{app:example}), and prompt templates (Sec.~\ref{app:prompts}).

\section{Related Work}
\label{sec:related}
Tool-using LLM agents that interleave reasoning with search, visit,
or system-level actions are now a mainstream
recipe~\citep{yao2023react,schick2023toolformerlanguagemodelsteach,qin2023toolllmfacilitatinglargelanguage,deng2023mind2web,OSWorld,yao2024taubenchbenchmarktoolagentuserinteraction},
and SFT on multi-turn trajectories from a stronger LLM is the
prevailing fine-tuning route~\citep{zeng-etal-2024-agenttuning}.

Closer in spirit, three lines of work pursue better SFT \emph{data}.
The ``less-is-more'' line picks SFT samples by hand, quality score,
or influence estimate, and shows that small selected corpora can
match larger noisy
ones~\citep{NEURIPS2023_ac662d74,cao2024instructionmininginstructiondata,liu2024makesgooddataalignment}.
Reasoning-step supervision edits or scores individual steps inside
one response, either by self-training new
rationales~\citep{zelikman2022starbootstrappingreasoningreasoning} or via step-level process
rewards~\citep{wang-etal-2024-math}. Trajectory synthesis
generates agentic training data from scratch via teacher-driven
agentic flows~\citep{mitra2024agentinstructgenerativeteachingagentic}.

These approaches operate across whole samples, inside a single
response, or by generating new data, leaving the internal
structure of an agent trajectory untouched. We
recast each trajectory as a round-level dependency DAG and apply
deterministic edits inside it, turning trajectory refinement into a
transparent graph-editing problem.

\section{Conclusion}
\label{sec:conclusion}
We recast each multi-turn agent trajectory as a round-level dependency DAG and apply three deterministic structural edits, with an optional LLM rephrasing variant. Across four multi-modal QA benchmarks, leaf prune (\textbf{L1}) anchors the high-accuracy end at $+1.7$\,pp over vanilla SFT, while strict merge with rephrasing (\textbf{L2a}) anchors the cost-efficient end by halving per-sample inference tokens at vanilla-SFT accuracy. The graph-editing view makes the structural assumptions behind each edit explicit and offers a general handle for refining multi-turn agent trajectories.

\section*{Limitations}
\label{sec:limitations}
Our study is scoped to one base model family (\texttt{Qwen3-VL-30B-A3B-Thinking}). The proposed edits act on training data rather than model weights, but the combined training and per-checkpoint evaluation already approaches our compute budget, so transfer across base models and modalities is left to follow-up work. Dependency annotation and merge-think rephrasing each add one annotator pass at data preparation time, and these one-time costs amortise quickly against the inference-time savings in Sec.~\ref{sec:main_results}.

\section*{Acknowledgments}
This work was supported by the Core Facility Platform of Computer Science and Communication, SIST, ShanghaiTech University.

\bibliography{custom}

\appendix

\newtcolorbox{strictpromptbox}[1][]{
  enhanced,
  colback=promptA!18, colframe=promptA!70!black, colbacktitle=promptA!75!black,
  coltitle=white, fonttitle=\small\bfseries,
  title={#1},
  boxrule=0.6pt, arc=2pt, left=4pt, right=4pt, top=2pt, bottom=2pt,
  fontupper=\scriptsize\ttfamily,
}
\newtcolorbox{mergepromptbox}[1][]{
  enhanced,
  colback=promptC!22, colframe=promptC!70!black, colbacktitle=promptC!72!black,
  coltitle=white, fonttitle=\small\bfseries,
  title={#1},
  boxrule=0.6pt, arc=2pt, left=4pt, right=4pt, top=2pt, bottom=2pt,
  fontupper=\scriptsize\ttfamily,
}
\newtcolorbox{judgepromptbox}[1][]{
  enhanced,
  colback=promptD!28, colframe=promptD!70!black, colbacktitle=promptD!78!black,
  coltitle=white, fonttitle=\small\bfseries,
  title={#1},
  boxrule=0.6pt, arc=2pt, left=4pt, right=4pt, top=2pt, bottom=2pt,
  fontupper=\scriptsize\ttfamily,
}
\newtcolorbox{cpathpromptbox}[1][]{
  enhanced,
  colback=promptB!18, colframe=promptB!70!black, colbacktitle=promptB!72!black,
  coltitle=white, fonttitle=\small\bfseries,
  title={#1},
  boxrule=0.6pt, arc=2pt, left=4pt, right=4pt, top=2pt, bottom=2pt,
  fontupper=\scriptsize\ttfamily,
}
\newtcolorbox{roundbox}[2][]{
  enhanced,
  colback=gray!3, colframe=gray!55!black, colbacktitle=gray!45!black,
  coltitle=white, fonttitle=\footnotesize\bfseries,
  title={#2},
  boxrule=0.4pt, arc=1.5pt, left=5pt, right=5pt, top=2pt, bottom=2pt,
  fontupper=\footnotesize, #1
}
\newtcolorbox{droppedroundbox}[2][]{
  enhanced,
  colback=warn!4, colframe=warn!55, colbacktitle=warn!65,
  coltitle=white, fonttitle=\footnotesize\bfseries,
  title={#2},
  boxrule=0.4pt, arc=1.5pt, left=5pt, right=5pt, top=2pt, bottom=2pt,
  fontupper=\footnotesize\color{warn!80!black}, #1
}
\newtcolorbox{mergedroundbox}[2][]{
  enhanced,
  colback=best!4, colframe=best!70!black, colbacktitle=best!80!black,
  coltitle=white, fonttitle=\footnotesize\bfseries,
  title={#2},
  boxrule=0.55pt, arc=1.5pt, left=5pt, right=5pt, top=2pt, bottom=2pt,
  fontupper=\footnotesize, #1
}
\newcommand{\rlabel}[1]{\textsf{\textbf{\footnotesize #1}}}
\newcommand{\rthink}{\rlabel{think}\hspace{0.6em}}
\newcommand{\rcall}{\rlabel{call}\hspace{0.95em}}
\newcommand{\rresp}{\rlabel{resp}\hspace{0.85em}}
\newcommand{\ranswer}{\rlabel{answer}\hspace{0.4em}}

\section{Training Setting}
\label{app:train}
All SFT systems (Vanilla SFT, Critical-Path, \textbf{L1}--\textbf{L3a}) fine-tune the
same \texttt{Qwen3-VL-30B-A3B-Thinking} base
model~\citep{bai2025qwen3vltechnicalreport} with full-parameter updates under identical
hyperparameters (Table~\ref{tab:train_setting}). Systems differ only
in their training trajectories. The training set is split 90/10 into
train/eval, and the best checkpoint per system is chosen by held-out
4-benchmark accuracy (Sec.~\ref{app:loss}). Implementation builds
on LLaMA-Factory~\citep{zheng-etal-2024-llamafactory}.

\begin{table}[htb]
\centering
\small
\setlength{\tabcolsep}{6pt}
\scalebox{0.95}{
\begin{tabular}{ll}
\toprule
Hyperparameter & Value \\
\midrule
Base model            & \texttt{Qwen3-VL-30B-A3B-Thinking} \\
Update type           & full-parameter \\
Epochs                & 4 \\
Learning rate         & $5\!\times\!10^{-6}$ \\
LR schedule           & cosine, 10\% warm-up \\
Precision             & bf16 \\
Optimizer sharding    & DeepSpeed ZeRO-3 \\
Gradient accumulation & 2 \\
\texttt{ddp\_timeout} & $1.8\!\times\!10^{8}$\,s \\
Train / eval split    & 90 / 10 \\
\bottomrule
\end{tabular}}
\caption{Training setting, shared across all SFT systems.}
\label{tab:train_setting}
\end{table}

\section{Inference Setting}
\label{app:inference}
We use a standard agent loop with one tool call per assistant turn and at most
128~rounds in total. Serving is via \texttt{vLLM} (TP$=8$, bf16). The
128-round limit acts as a safety net. We report \emph{stuck rate} as
the fraction of evaluation samples that hit this limit (see
Sec.~\ref{sec:analysis} and Sec.~\ref{app:stuck}).

\section{Per-Dataset Inference Cost}
\label{app:detailed_perf}
Tables~\ref{tab:perf_msgs}--\ref{tab:perf_toks} expand the
per-sample message and token columns of Table~\ref{tab:main} into a
system~$\times$~dataset grid. Both metrics are averaged across the
samples of each benchmark, and the rightmost column reproduces the
4-benchmark average reported in the main table.
\textbf{L2a} is uniformly the cheapest in messages on every benchmark (Table~\ref{tab:perf_msgs}). Among SFT systems it is the best on all four benchmarks in tokens (Table~\ref{tab:perf_toks}). Zero-shot's apparently low token counts on LiveVQA, HLE, and MMSearch, and its 119.26 messages on HLE, are a side-effect of the stuck-loop rate of 45.5\% on HLE (Sec.~\ref{app:stuck}), which produces short repeated messages until the round limit.

\begin{table}[tb]
\centering
\small
\setlength{\tabcolsep}{3pt}
\begin{tabular}{lccccc}
\toprule
System & \makecell{Simple\\VQA} & \makecell{Live\\VQA} & HLE
       & \makecell{MM\\Search} & \textbf{Avg.} \\
\midrule
Zero-shot     & 16.69 & 18.71 & \warntext{119.26}\sub{\dag} & 12.03 & 41.67 \\
Vanilla SFT   & 12.51 & 26.31 & 36.60 & 20.33 & 23.94 \\
Critical-Path & 19.27 & 25.79 & 43.41 & 20.87 & 27.33 \\
\midrule
\textbf{L1}   & 11.09 & 15.37 & 21.58 & 15.56 & 15.90 \\
\textbf{L2}   & 14.03 & 17.34 & 36.53 & 21.11 & 22.25 \\
\textbf{L3}   & 26.09 & 28.56 & 39.90 & 26.20 & 30.19 \\
\textbf{L2a}  & \best{10.84} & \best{14.93} & \best{20.04} & \best{11.39} & \best{14.30} \\
\textbf{L3a}  & 11.25 & 15.62 & 27.61 & 12.04 & 16.63 \\
\bottomrule
\end{tabular}
\caption{Per-sample average \textbf{message count} by system and
dataset. \best{Bold} = best in column (among SFT systems).
\sub{\dag}\,Zero-shot artefact, see prose above.}
\label{tab:perf_msgs}
\end{table}

\begin{table}[tb]
\centering
\small
\setlength{\tabcolsep}{3pt}
\begin{tabular}{lccccc}
\toprule
System & \makecell{Simple\\VQA} & \makecell{Live\\VQA} & HLE
       & \makecell{MM\\Search} & \textbf{Avg.} \\
\midrule
Zero-shot     & 8{,}278 & \best{10{,}757}\sub{\dag} & \best{11{,}488}\sub{\dag} & \best{5{,}385} & \best{8{,}977}\sub{\dag} \\
Vanilla SFT   & 10{,}106 & 28{,}176 & 26{,}880 & 18{,}596 & 20{,}940 \\
Critical-Path & 12{,}531 & 22{,}804 & 26{,}820 & 16{,}005 & 19{,}540 \\
\midrule
\textbf{L1}   & 8{,}241 & 18{,}081 & 16{,}012 & 16{,}293 & 14{,}657 \\
\textbf{L2}   & 9{,}133 & 14{,}461 & 21{,}750 & 18{,}018 & 15{,}841 \\
\textbf{L3}   & 25{,}532 & 39{,}353 & 30{,}834 & 31{,}396 & 31{,}779 \\
\textbf{L2a}  & \textbf{7{,}665} & 13{,}264 & 13{,}816 & 9{,}161 & 10{,}977 \\
\textbf{L3a}  & 8{,}485 & 14{,}435 & 18{,}709 & 13{,}437 & 13{,}766 \\
\bottomrule
\end{tabular}
\caption{Per-sample average \textbf{token count} by system and
dataset, computed via \texttt{Qwen2Tokenizer} on the full
assistant+user trajectory. \best{Bold} = best in column.
\sub{\dag}\,Zero-shot artefact, see prose above.}
\label{tab:perf_toks}
\end{table}

\begin{table}[tb]
\centering
\small
\setlength{\tabcolsep}{3pt}
\begin{tabular}{lcccc}
\toprule
& \makecell{Avg.\\Acc.\,(\%)} & \makecell{Avg.\\\#msgs}
& \makecell{Acc/msg}
& \makecell{Acc/tok\\$\times 10^{3}$} \\
\midrule
Zero-shot     & 44.68 & 41.67 & 1.07 & 4.98\sub{\dag} \\
Vanilla SFT   & 48.35 & 23.94 & 2.02 & 2.31 \\
Critical-Path & 44.32 & 27.33 & 1.62 & 2.27 \\
\midrule
\textbf{L1}   & \best{50.04} & 15.90 & 3.15 & 3.41 \\
\textbf{L2}   & 48.22 & 22.25 & 2.17 & 3.04 \\
\textbf{L3}   & 48.07 & 30.19 & 1.59 & 1.51 \\
\textbf{L2a}  & 48.10 & \best{14.30} & \best{3.36} & \best{4.38} \\
\textbf{L3a}  & 47.55 & 16.63 & 2.86 & 3.45 \\
\bottomrule
\end{tabular}
\caption{Cost-effectiveness summary, restating the bottom block of
Table~\ref{tab:main}. \best{Bold} = best in column. Both
cost-effectiveness ratios are computed from the row's average
accuracy and the corresponding average message / token count.
\sub{\dag}\,Zero-shot's Acc/tok ratio inherits the stuck-loop
artefact in Table~\ref{tab:perf_toks}. \textbf{L2a} strictly beats every SFT baseline on both ratios.}
\label{tab:perf_costeff}
\end{table}

\section{Stuck-Rate Note}
\label{app:stuck}
The per-dataset stuck-rate breakdown is given in Sec.~\ref{sec:analysis}. We note two side
observations not covered there. First, the $45.5\%$ stuck rate of
zero-shot on HLE explains the anomalously low average-token figure of
zero-shot in Sec.~\ref{app:detailed_perf}, since stuck rollouts
end without an \texttt{<answer>} and are truncated at the 128-round
ceiling. Second, the residual elevation of \textbf{L3a} on HLE (versus \textbf{L2a})
is consistent with the format-mismatch interpretation in
Sec.~\ref{sec:analysis}, since merge-think rephrasing cannot fully
reshape the multi-response \texttt{<tool\_response>} block that \textbf{L3}
produces.

\section{Merge Granularity and Inference-Time Format Alignment}
\label{app:merge_format}

\textbf{L2}/\textbf{L2a} and \textbf{L3}/\textbf{L3a} use closely related merge primitives but show
distinctly different inference behaviour
(Tables~\ref{tab:perf_msgs}--\ref{tab:perf_toks},
Sec.~\ref{app:stuck}). This appendix explains where the gap
arises and states the design rule it suggests.

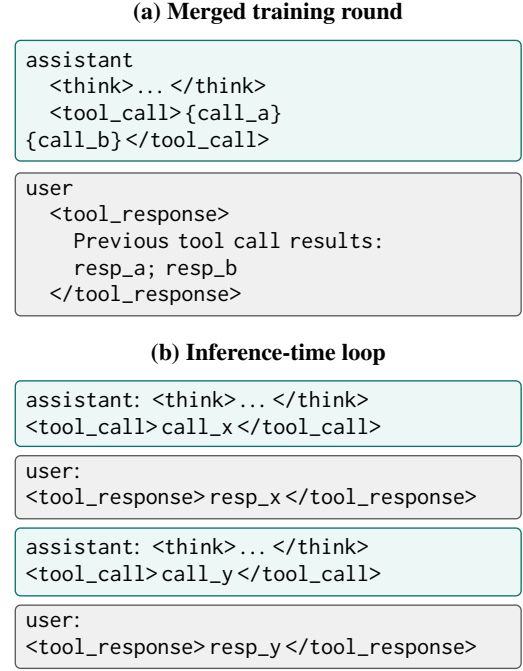
\begin{figure}[tb]
\centering
\begin{tikzpicture}[
  every node/.style={font=\footnotesize},
  asst/.style={rectangle,draw=best!75!black,fill=best!8,
               rounded corners=2pt,inner sep=4pt,align=left,
               text width=6.4cm,line width=0.5pt},
  usr/.style ={rectangle,draw=gray!70!black,fill=gray!12,
               rounded corners=2pt,inner sep=4pt,align=left,
               text width=6.4cm,line width=0.5pt},
  ttl/.style ={font=\small\bfseries,anchor=south},
]

\node[ttl] (ta) at (0,0) {(a) Merged training round};
\node[asst,anchor=north,below=2pt of ta] (a1) {%
  \texttt{assistant}\\
  \quad\texttt{<think>}\,\dots\,\texttt{</think>}\\
  \quad\texttt{<tool\_call>}\,\texttt{\{call\_a\}\ \{call\_b\}}\,\texttt{</tool\_call>}};
\node[usr,anchor=north,below=4pt of a1] (u1) {%
  \texttt{user}\\
  \quad\texttt{<tool\_response>}\\
  \quad\quad\texttt{Previous tool call results:}\\
  \quad\quad\texttt{resp\_a; resp\_b}\\
  \quad\texttt{</tool\_response>}};

\node[ttl] (tb) at (0,-4.5) {(b) Inference-time loop};
\node[asst,anchor=north,below=2pt of tb] (a2) {%
  \texttt{assistant}: \texttt{<think>}\,\dots\,\texttt{</think>} \texttt{<tool\_call>}\,\texttt{call\_x}\,\texttt{</tool\_call>}};
\node[usr,anchor=north,below=3pt of a2] (u2) {%
  \texttt{user}: \texttt{<tool\_response>}\,\texttt{resp\_x}\,\texttt{</tool\_response>}};
\node[asst,anchor=north,below=3pt of u2] (a3) {%
  \texttt{assistant}: \texttt{<think>}\,\dots\,\texttt{</think>} \texttt{<tool\_call>}\,\texttt{call\_y}\,\texttt{</tool\_call>}};
\node[usr,anchor=north,below=3pt of a3] (u3) {%
  \texttt{user}: \texttt{<tool\_response>}\,\texttt{resp\_y}\,\texttt{</tool\_response>}};

\end{tikzpicture}
\caption{Message pattern of a merged training round (a) vs.\ a
typical inference-time loop (b). In (a) a single \texttt{<tool\_call>}
tag holds $k$ JSON calls and a single \texttt{<tool\_response>} tag
holds the $k$ corresponding results, whereas the loop in (b) usually
emits one call/result per turn.}
\label{fig:format_gap}
\end{figure}

\paragraph{The Format Gap.} The inference agent loop is strictly
serial. In the great majority of assistant turns, one \texttt{<tool\_call>} tag
carries a single JSON call, and the next user turn returns one
\texttt{<tool\_response>} with that call's result. Merging preserves
the canonical outer tags on both sides, but concatenates the
$k\!\geq\!2$ siblings' \emph{bodies} inside them. The merged
\texttt{<tool\_call>} now contains $k$ JSON calls, and the merged
\texttt{<tool\_response>}, prefixed
``Previous tool call results:'', contains the $k$
corresponding results. A round whose two outer tags hold $k$ items
each is a configuration the inference loop does not emit
(Fig.~\ref{fig:format_gap}).

\paragraph{Why L2 and L3 Differ.}
Both edits create the multi-item pattern above, but at different rates and with different downstream structure
(Table~\ref{tab:merge_counts}).
\emph{(i)~Frequency.} \textbf{L2} requires matching parents \emph{and} children and yields $1{,}044$ fusions;
\textbf{L3} matches parents only and yields $3{,}234$ (about $3\times$), mostly as more merges per sample
(samples affected rise only $19.0\%$\,$\to$\,$21.3\%$).
\emph{(ii)~Downstream alignment.} \textbf{L2} merges only siblings that feed the same next round, so each
merged user turn still has a single consumer---the same ``one user turn $\to$ one next \texttt{<think>}''
pattern as inference.
\textbf{L3} can bundle siblings that originally fed different children, gaining denser compression at the
cost of that one-to-one structure; keeping the stricter \textbf{L2} rule is the more inference-aligned choice.

\begin{table}[h]
\centering
\small
\setlength{\tabcolsep}{8pt}
\begin{tabular}{lcc}
\toprule
Merge group size & \textbf{L2} & \textbf{L3} \\
\midrule
2               &  709 & 2{,}491 \\
3               &  230 &    586 \\
4               &   83 &    122 \\
$\geq 5$        &   22 &     35 \\
\midrule
Total fusions    & 1{,}044 & 3{,}234 \\
Samples affected & 1{,}175\,(19.0\%) & 1{,}322\,(21.3\%) \\
\bottomrule
\end{tabular}
\caption{Merge groups by size and fraction of training samples
affected, on the 6{,}196-trajectory corpus.}
\label{tab:merge_counts}
\end{table}

\paragraph{Design Rule.} A round-level merge should preserve the
downstream consumption pattern of the original trajectory.
Concretely, require the merge criterion to match on both parents
\emph{and} children, and prefer leaf-pruning over fusion when
downstream consumers differ. This keeps the training-time user-turn
distribution close to what the inference loop produces.

\begin{table*}[t]
  \centering
  \scriptsize
  \setlength{\tabcolsep}{1.8pt}
  \begin{tabular}{@{}cl c|c|cc|cc|cc|cc|cc@{}}
  \toprule
  & & \makecell{Vanilla\\SFT}
  & +DPO[\textbf{CP}]
  & +DPO[\textbf{L1}]  & $\Delta_V$
  & +DPO[\textbf{L2}]  & $\Delta_V$
  & +DPO[\textbf{L3}]  & $\Delta_V$
  & +DPO[\textbf{L2a}] & $\Delta_V$
  & +DPO[\textbf{L3a}] & $\Delta_V$ \\
  \midrule
  \multirow{5}{*}{\emph{\makecell[c]{Task\\acc.}}}
   & SimpleVQA  & 68.67 & \textit{diverged}$^{\dagger}$ & 70.00 & \dpos{+1.3} & \best{70.67} & \dpos{+2.0} & 67.33 & \dneg{-1.3} & 68.33 & \dneg{-0.3} & 66.67 & \dneg{-2.0} \\
   & LiveVQA    & 51.00 & \textit{diverged}$^{\dagger}$ & 52.00 & \dpos{+1.0} & \best{53.67} & \dpos{+2.7} & 50.67 & \dneg{-0.3} & 50.33 & \dneg{-0.7} & 50.00 & \dneg{-1.0} \\
   & HLE        &  9.39 & \textit{diverged}$^{\dagger}$ & 12.42 & \dpos{+3.0} & 12.12 & \dpos{+2.7} & \best{12.73} & \dpos{+3.3} & 12.42 & \dpos{+3.0} & 10.91 & \dpos{+1.5} \\
   & MMSearch   & \best{64.33} & \textit{diverged}$^{\dagger}$ & 54.39 & \dneg{-9.9} & 59.06 & \dneg{-5.3} & 57.89 & \dneg{-6.4} & 53.22 & \dneg{-11.1} & 60.23 & \dneg{-4.1} \\
   & \textbf{Avg.\ Acc.} & 48.35 & \textit{diverged}$^{\dagger}$ & 47.20 & \dneg{-1.2} & \best{48.88} & \dpos{+0.5} & 47.16 & \dneg{-1.2} & 46.08 & \dneg{-2.3} & 46.95 & \dneg{-1.4} \\
  \midrule
  \multirow{2}{*}{\emph{\makecell[c]{Inf.\\eff.}}}
   & Avg.\ \#msgs & 23.94 & \textit{diverged}$^{\dagger}$ & 11.60 & \dpos{-12.3} & 11.43 & \dpos{-12.5} & 12.03 & \dpos{-11.9} & 11.25 & \dpos{-12.7} & \best{10.23} & \dpos{-13.7} \\
   & \makecell[l]{Avg.\ \#toks \tiny ($10^{3}$)} & 20.9 & \textit{diverged}$^{\dagger}$ & 4.8 & \dpos{-16.1} & 4.5 & \dpos{-16.5} & 4.7 & \dpos{-16.2} & 4.6 & \dpos{-16.4} & \best{4.0} & \dpos{-16.9} \\
  \midrule
  \multirow{2}{*}{\emph{\makecell[c]{Cost-\\eff.}}}
   & Acc/msg & 2.02 & \textit{diverged}$^{\dagger}$ & 4.07 & \dpos{+2.0} & 4.28 & \dpos{+2.3} & 3.92 & \dpos{+1.9} & 4.09 & \dpos{+2.1} & \best{4.59} & \dpos{+2.6} \\
   & Acc/$10^{3}$tok & 2.31 & \textit{diverged}$^{\dagger}$ & 9.85 & \dpos{+7.5} & 10.98 & \dpos{+8.7} & 9.99 & \dpos{+7.7} & 10.10 & \dpos{+7.8} & \best{11.63} & \dpos{+9.3} \\
  \bottomrule
  \end{tabular}
  \caption{Shared-base DPO from Vanilla SFT. $\Delta_V$: +DPO minus Vanilla SFT. $^{\dagger}$Diverged (context overflow).}
  \label{tab:dpo_transfer}
\end{table*}

\begin{table*}[t]
  \centering
  \scriptsize
  \setlength{\tabcolsep}{1.8pt}
  \begin{tabular}{@{}cl ccc|ccc|ccc|ccc|ccc|ccc@{}}
  \toprule
  & & \multicolumn{3}{c|}{\textbf{Critical-P.}}
    & \multicolumn{3}{c|}{\textbf{L1}}
    & \multicolumn{3}{c|}{\textbf{L2}}
    & \multicolumn{3}{c|}{\textbf{L3}}
    & \multicolumn{3}{c|}{\textbf{L2a}}
    & \multicolumn{3}{c}{\textbf{L3a}} \\
  & & SFT & +DPO & $\Delta$
    & SFT & +DPO & $\Delta$
    & SFT & +DPO & $\Delta$
    & SFT & +DPO & $\Delta$
    & SFT & +DPO & $\Delta$
    & SFT & +DPO & $\Delta$ \\
  \midrule
  \multirow{5}{*}{\emph{\makecell[c]{Task\\acc.}}}
   & SimpleVQA & 60.67 & 68.00 & \dpos{+7.3}  & \best{70.67} & 68.00 & \dneg{-2.7} & 68.67 & 64.67 & \dneg{-4.0} & 66.67 & 66.67 & $0$           & 67.33 & 68.67 & \dpos{+1.3} & 64.33 & 66.00 & \dpos{+1.7} \\
   & LiveVQA   & 46.33 & 48.67 & \dpos{+2.3}  & \best{53.67} & 51.00 & \dneg{-2.7} & 51.33 & 49.33 & \dneg{-2.0} & 50.33 & 51.00 & \dpos{+0.7}  & 51.00 & 50.67 & \dneg{-0.3} & 51.00 & 52.67 & \dpos{+1.7} \\
   & HLE       & 11.21 &  8.48 & \dneg{-2.7}  & 10.30 & 10.91 & \dpos{+0.6} & 10.30 & 12.12 & \dpos{+1.8} & 11.52 & 13.33 & \dpos{+1.8}  & 11.52 & \best{14.55} & \dpos{+3.0} & 8.79 & 10.91 & \dpos{+2.1} \\
   & MMSearch  & 59.06 & 58.48 & \dneg{-0.6}  & 65.50 & 62.57 & \dneg{-2.9} & 62.57 & 60.82 & \dneg{-1.8} & 63.74 & \best{66.08} & \dpos{+2.3}  & 62.57 & 59.06 & \dneg{-3.5} & 66.08 & 63.74 & \dneg{-2.3} \\
   & \textbf{Avg.\ Acc.} & 44.32 & 45.91 & \dpos{+1.6}  & \best{50.04} & 48.12 & \dneg{-1.9} & 48.22 & 46.73 & \dneg{-1.5} & 48.07 & 49.27 & \dpos{+1.2} & 48.10 & 48.24 & \dpos{+0.1} & 47.55 & 48.33 & \dpos{+0.8} \\
  \midrule
  \multirow{2}{*}{\emph{\makecell[c]{Inf.\\eff.}}}
   & Avg.\ \#msgs & 27.33 & 14.12 & \dpos{-13.2} & 15.90 & \best{13.86} & \dpos{-2.0} & 22.25 & 16.52 & \dpos{-5.7} & 30.19 & 22.08 & \dpos{-8.1} & 14.30 & 14.48 & \dneg{+0.2} & 16.63 & 15.60 & \dpos{-1.0} \\
   & \makecell[l]{Avg.\ \#toks \tiny ($10^{3}$)} & 19.5 & 5.2 & \dpos{-14.4} & 14.7 & \best{5.3} & \dpos{-9.4} & 15.8 & 6.1 & \dpos{-9.8} & 31.8 & 8.8 & \dpos{-23.0} & 11.0 & 5.5 & \dpos{-5.5} & 13.8 & 5.8 & \dpos{-7.9} \\
  \midrule
  \multirow{2}{*}{\emph{\makecell[c]{Cost-\\eff.}}}
   & Acc/msg & 1.62 & 3.25 & \dpos{+1.6} & 3.15 & 3.47 & \dpos{+0.3} & 2.17 & 2.83 & \dpos{+0.7} & 1.59 & 2.23 & \dpos{+0.6} & 3.36 & 3.33 & \dneg{-0.0} & 2.86 & 3.10 & \dpos{+0.2} \\
   & Acc/$10^{3}$tok & 2.27 & 8.90 & \dpos{+6.6} & 3.41 & \best{9.14} & \dpos{+5.7} & 3.04 & 7.71 & \dpos{+4.7} & 1.51 & 5.62 & \dpos{+4.1} & 4.38 & 8.80 & \dpos{+4.4} & 3.45 & 8.28 & \dpos{+4.8} \\
  \bottomrule
  \end{tabular}
  \caption{Matched-base DPO on each setting's own SFT checkpoint. $\Delta$: +DPO minus that setting's SFT.}
  \label{tab:dpo_main}
\end{table*}

\section{DPO on Refined-vs-Raw Preference Pairs}
\label{app:dpo}

Each structural edit yields a free preference pair that shares the final answer, taking the refined trajectory as ``chosen'' and the raw trajectory as ``rejected'' (counts: Critical-Path $4{,}446$; \textbf{L1} $986$; \textbf{L2}/\textbf{L2a} $1{,}175$; \textbf{L3}/\textbf{L3a} $1{,}322$). We plug these pairs into a follow-up DPO~\citep{rafailov2024directpreferenceoptimizationlanguage} stage in two recipes that differ only in the DPO starting checkpoint. 

\paragraph{Shared-Base Recipe.} Train one Vanilla SFT model on raw trajectories, then run DPO from that shared base while swapping each setting's preference pairs, Table~\ref{tab:dpo_transfer}. All five transferable settings collapse inference cost far below Vanilla SFT. \textbf{L2} pairs are strongest on accuracy, \textbf{L3a} pairs on cost-efficiency. Critical-Path pairs push Vanilla SFT off distribution and the run diverges.

\paragraph{Matched-Base Recipe.} Start DPO from each setting's own SFT checkpoint, Table~\ref{tab:dpo_main}. Accuracy is better preserved than under the shared base, while tokens still drop, though they remain above the shared-base band. Shared-base is stronger for cost-efficient deployment. Matched-base is safer when raw accuracy is the priority.

\section{DPO Setting}
\label{app:dpo_setting}

We train with sigmoid DPO plus an NLL-on-chosen anchor (Table~\ref{tab:dpo_setting}). \textbf{L1} uses $200$ steps and $\alpha_{\mathrm{RPO}}=0.7$; others use $300$ steps and $\alpha_{\mathrm{RPO}}=0.5$. Checkpoints every $50$ steps are selected by mix-dev accuracy rather than eval-loss.

\begin{table}[t]
\centering
\small
\setlength{\tabcolsep}{3pt}
\scalebox{0.8}{
\begin{tabular}{@{}ll@{}}
\toprule
Hyperparameter & Value \\
\midrule
Loss & sigmoid DPO + NLL-on-chosen \\
$\beta$ (DPO temp.) & 0.1 \\
$\alpha_{\mathrm{RPO}}$ (NLL wt.) & 0.5 (0.7 for \textbf{L1}) \\
Max steps & 300 (200 for \textbf{L1}) \\
Learning rate & $5\!\times\!10^{-7}$ \\
LR schedule & cosine, 5\% warm-up \\
Max seq.\ length & 16{,}384 \\
Per-device batch & 1 \\
Grad.\ accum. & 2 \\
Global batch & 16 \\
Hardware & $8\times80$\,GiB GPUs \\
Optimiser & DeepSpeed ZeRO-3 \\
Attention & SDPA \\
\bottomrule
\end{tabular}
}
\caption{DPO training setting (shared across SFT bases).}
\label{tab:dpo_setting}
\end{table}

\begin{figure*}[t]
\centering
\includegraphics[width=0.82\linewidth]{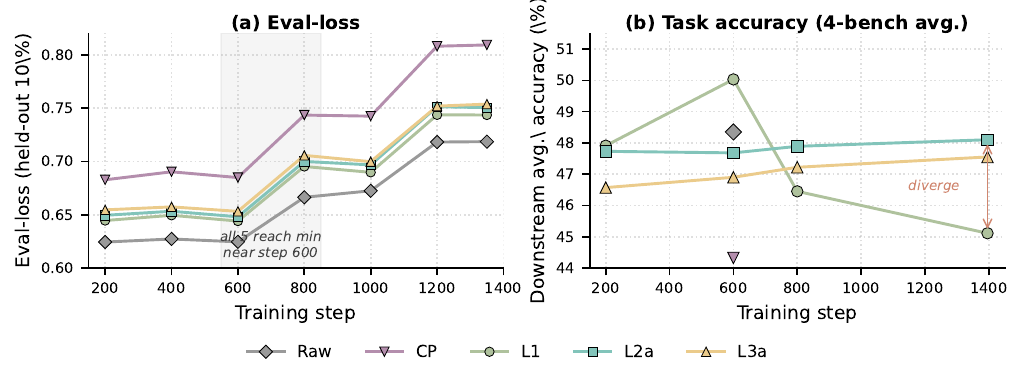}
\caption{Eval-loss curves (left, dense schedule) vs.\ downstream
4-benchmark accuracy (right, four evaluated checkpoints) for \textbf{L1},
\textbf{L2a}, \textbf{L3a}.}
\label{fig:loss_vs_acc}
\end{figure*}

\section{Checkpoint Selection}
\label{app:loss}

Eval-loss on the held-out 10\% of the training data was recorded at
every 200 steps (Table~\ref{tab:loss_vs_acc}, top half, and
Fig.~\ref{fig:loss_vs_acc}~left). Downstream task accuracy,
however, requires running the full agent loop on the four
benchmarks for every candidate checkpoint, which is expensive.
Within our compute budget we evaluated four checkpoints for our
edits, namely steps~200, 600, 800, and 1396 (final), and a single
checkpoint at step~600 for Vanilla~SFT and Critical-Path (selected
by the same rule applied to our edits). A denser sweep is left to
follow-up work.

\begin{table}[h]
\centering
\small
\setlength{\tabcolsep}{5pt}
\begin{tabular}{lccccc}
\toprule
Step & Raw & CP & \textbf{L1} & \textbf{L2a} & \textbf{L3a} \\
\midrule
\multicolumn{6}{l}{\emph{eval-loss (held-out 10\% of training data)}} \\
200  & 0.6243 & 0.6827 & 0.6445 & 0.6494 & 0.6545 \\
400  & 0.6272 & 0.6902 & 0.6494 & 0.6532 & 0.6573 \\
600  & 0.6245 & 0.6848 & 0.6440 & 0.6480 & 0.6531 \\
800  & 0.6663 & 0.7434 & 0.6952 & 0.6999 & 0.7056 \\
1000 & 0.6724 & 0.7424 & 0.6897 & 0.6967 & 0.6997 \\
1200 & 0.7181 & 0.8080 & 0.7436 & 0.7512 & 0.7518 \\
1350 & 0.7185 & 0.8092 & 0.7435 & 0.7504 & 0.7537 \\
\midrule
\multicolumn{6}{l}{\emph{downstream task accuracy (\%, 4-bench avg.)}} \\
200  & --- & --- & 47.91 & 47.73 & 46.57 \\
600  & 48.35 & 44.32 & \best{50.04} & 47.68 & 46.90 \\
800  & --- & --- & 46.45 & 47.89 & 47.22 \\
1396 & --- & --- & 45.11 & \best{48.10} & \best{47.55} \\
\bottomrule
\end{tabular}
\caption{Eval-loss vs.\ downstream accuracy.}
\label{tab:loss_vs_acc}
\end{table}

\paragraph{Best-Checkpoint Summary.} By 4-benchmark average
accuracy, \textbf{L1, Vanilla SFT, and Critical-Path} pick
step~\textbf{600}, while \textbf{L2a and \textbf{L3a}} actually pick
step~\textbf{1396}.

\paragraph{Eval-Loss Decouples from Accuracy.} All three of \textbf{L1},
\textbf{L2a}, \textbf{L3a} follow a similar eval-loss trajectory (minimum near
step~600, then rising), but downstream accuracy diverges.
\textbf{L1} peaks at step~600 ($50.04\%$), drops to $46.45\%$ at step~800,
and reaches $45.11\%$ at step~1396 ($-4.9$\,pp from peak). \textbf{L2a} and
\textbf{L3a} instead climb \emph{monotonically} from step~600 onwards
($47.68 \!\to\! 47.89 \!\to\! 48.10$ for \textbf{L2a}, and $46.90 \!\to\!
47.22 \!\to\! 47.55$ for \textbf{L3a}, all in \%). Picking the checkpoint by
eval-loss alone would have chosen step~600 for everyone, which is
exactly \textbf{L1}'s peak but a poor choice for \textbf{L2a}/\textbf{L3a}.
Practitioners training on merged trajectories should therefore
select checkpoints by downstream metrics rather than eval-loss.
Continuing training beyond step~1396 would extend the 4-epoch SFT
budget and require evaluating additional checkpoints on the full
agent loop, which doubles inference cost per added checkpoint.
Eval-loss has been rising since step~600, so further training
carries a non-trivial overfitting risk despite the still-climbing
downstream metric for \textbf{L2a}/\textbf{L3a}. A larger compute-budget
sweep is left to follow-up work.

\clearpage
\onecolumn
\section{Qualitative Example}
\label{app:example}

\paragraph{Source Sample.} Training-corpus sample \texttt{\_idx=1581}.
The user shows an image carrying ``SAGE Publishing'' branding and a
``Live Product Training'' banner, and asks: \emph{``Who founded SAGE
Publishing, and within the first decade after that, who established
the company's first international office in London?''} The gold
answer ends with ``Sara Miller McCune (with George McCune) founded
SAGE in 1965, and SAGE Publications Ltd.\ was established in London
in 1971''.

In what follows, each grey box is one round of the original
trajectory, and the bar at the top names the round and its role.
Round~1 is the user question (image~+~text). Rounds~2--4 are
assistant turns whose \rlabel{think} and \rlabel{call} are followed
by a user-side \rlabel{resp}. Round~5 is the final assistant
\rlabel{answer}. Red-bordered boxes mark rounds removed by an
edit, and teal-bordered boxes mark rounds produced by merging.

\subsection*{Original Trajectory (8 Messages, 5 Rounds)}

\begin{figure}[!htb]
\begin{roundbox}{R1 \textnormal{(user)}}
{[image: SAGE Publishing logo on a Live Product Training banner]}\\
Who founded SAGE Publishing, and within the first decade after
that, who established the company's first international office in
London?
\end{roundbox}
\vspace{2pt}
\begin{roundbox}{R2 \textnormal{(assistant + user)}}
\rthink Identify who founded SAGE Publishing, then look up the
London office. Start with the founder.\\
\rcall \texttt{web\_search("Who founded SAGE Publishing")}\\
\rresp ``24-year-old Sara Miller McCune founded Sage in 1965 \dots''
(Wikipedia / About-SAGE)
\end{roundbox}
\vspace{2pt}
\begin{roundbox}{R3 \textnormal{(assistant + user)}}
\rthink Now find who set up SAGE's first international office in
London.\\
\rcall \texttt{web\_search("SAGE first international office London established by who and when")}\\
\rresp \emph{Same} About-SAGE snippet as R2, with no new fact about
the London office.
\end{roundbox}
\vspace{2pt}
\begin{roundbox}{R4 \textnormal{(assistant + user)}}
\rthink The previous result added nothing new. Try SAGE's own
50-year history page directly.\\
\rcall \texttt{visit\_url("https://us.sagepub.com/.../sage-50-year-history.pdf")}\\
\rresp ``1971: SAGE Publications, Ltd.\ established in London. Sara
and George hired David Brooks \dots''
\end{roundbox}
\vspace{2pt}
\begin{roundbox}{R5 \textnormal{(assistant)}}
\rthink I have what I need.\\
\ranswer Sara Miller McCune (with George McCune) founded SAGE, and
the first international office was established in London in 1971.
\end{roundbox}
\end{figure}

\subsection*{Annotated DAG (Prompt~A)}

Running the strict prompt (Sec.~\ref{app:prompts}, Prompt~A) on this
trajectory, the annotator LLM emits three edges with the following
abridged reasons.
\begin{itemize}[leftmargin=12pt,itemsep=1pt,topsep=2pt]
\item \texttt{R2$\to$R3}: ``Round~2 produced the concrete founder
identification \emph{Sara Miller McCune}, which Round~3 uses to
frame the next search.''
\item \texttt{R4$\to$R5}: ``Round~4 produced the key historical fact
that \emph{SAGE Publications, Ltd.\ was established in London in
1971}, which Round~5 uses to conclude the first international
office was in London within the first decade.''
\item \texttt{R2$\to$R5}: ``Round~2 established Sara Miller McCune
as the founder, which Round~5 combines with the London-1971 fact to
answer the full question.''
\end{itemize}

\paragraph{No Edge Originates at R1.} Prompt~A asks the annotator to
mark only globally load-bearing edges, where round~$j$ uses a
specific retrieved fact that round~$i$ produced. The user query
itself (R1) is the topic of every later round but contributes no
\emph{new retrieved fact} that a downstream round consumes, so the
strict criterion correctly omits all \texttt{R1$\to$*} edges. Two
structural guarantees keep this safe-by-construction. First, leaf
pruning removes only nodes with out-degree~0 that are neither
round~1 nor the highest-indexed round, so R1 is never a deletion
candidate even when it has no edges in the annotated DAG. Second, the merge function (Sec.~\ref{sec:rephrase}) skips any
merge group whose smallest round index is below~2, so R1's message
is left untouched even if it appeared in a merged set. As a result
the merge topology is identical whether or not the annotator emits
\texttt{R1$\to$*} edges.

The resulting DAG over annotated edges is
\texttt{R2$\to$R3},\, \texttt{R2$\to$R5}, and \texttt{R4$\to$R5}.
R3 has out-degree~0 (no downstream round uses anything it produced),
which marks it as the L1 leaf.

\subsection*{After \textbf{L1} (Leaf Prune): 6 Messages, 4 Rounds}

R3 has out-degree~0 in the annotated DAG, so \textbf{L1} drops it.
R1, R2, R4, R5 remain unchanged.

\begin{figure}[!htb]
\begin{droppedroundbox}{R3 \textnormal{(pruned by \textbf{L1}: out-degree 0)}}
\rthink \emph{(removed)}\quad
\rcall \emph{(removed)}\quad
\rresp \emph{(removed)}
\end{droppedroundbox}
\end{figure}

\noindent The kept trajectory after \textbf{L1} is \texttt{R1 $\to$ R2 $\to$
R4 $\to$ R5}, i.e.\ the original boxes for R1, R2, R4, R5 unchanged
(not redrawn).

\subsection*{After \textbf{L2a} (Strict Merge + LLM Rephrase): 4 Messages, 3 Rounds}

In the annotated DAG after \textbf{L1}, R2 and R4 both have an empty
parent set (Prompt~A emits no \texttt{R1$\to$*} edge) and the same
child set ($\{\text{R5}\}$). The strict-merge rule fuses them into
one round \texttt{R2+4}. Note that even if Prompt~A had emitted
\texttt{R1$\to$R2} and \texttt{R1$\to$R4}, both rounds would then
share parents $\{\text{R1}\}$ and child $\{\text{R5}\}$, and the
same merge would fire. Their two \rlabel{think} bodies are
concatenated and rephrased by the annotator LLM into a single
coherent passage (the \texttt{<think>} shown below). The two
original calls are emitted as two JSON arguments inside a single
\texttt{<tool\_call>} tag, and the two \rlabel{resp} bodies are
prefixed with ``\textsf{Previous tool call results:}'' and emitted
in one user turn.

\begin{figure}[!htb]
\begin{mergedroundbox}{R2+4 \textnormal{(merged by \textbf{L2a}, covering original R2 and R4)}}
\rthink The image shows SAGE Publishing's branding on a Live
Product Training banner. The question has two parts (who founded
SAGE Publishing and who established its first London office within
the first decade) and these are likely connected, so first locate
the founder, then look at SAGE's early history for when and how the
London office was set up.\\
\rcall \texttt{web\_search("Who founded SAGE Publishing"),}\\
\hspace*{2.4em}\texttt{visit\_url("https://us.sagepub.com/.../sage-50-year-history.pdf")}\\
\rresp \textsf{Previous tool call results:}\\
\hspace*{1.2em}(1) ``24-year-old Sara Miller McCune founded Sage in
1965 \dots''\\
\hspace*{1.2em}(2) ``1971: SAGE Publications, Ltd.\ established in
London. Sara and George hired David Brooks \dots''
\end{mergedroundbox}
\end{figure}

\noindent The full \textbf{L2a} trajectory is \texttt{R1 $\to$ R2+4 $\to$
R5} (R5 unchanged from the original).

\paragraph{Take-Aways.} Two distinct effects compose on the same
trajectory. First, Leaf~Prune (\textbf{L1}) removes the redundant
follow-up search whose evidence was already in R2. Second,
Strict Merge with rephrase (\textbf{L2a}) folds the two
substantively different sub-queries (``who founded'' vs.\ ``where
was the first international
office'') into one round of reasoning whose \texttt{<think>} reads as
a single chain of thought, not two disjoint segments concatenated with
a semicolon. The final answer in R5 is unchanged.

\subsection*{Two Further DAG-Only Cases}

To isolate each edit's topological action, we show two more samples
from the training corpus as bare DAGs, omitting the per-round content.
The first sample triggers only L1, the second only L3.

\paragraph{Sample~A (\texttt{\_idx=6061}), L1 Only.}
The annotated DAG has edges \texttt{R1$\to$R2}, \texttt{R1$\to$R3},
\texttt{R3$\to$R4}. Round~R2's out-degree is~0 and it is not the
highest-indexed round, so \textbf{L1} prunes it. L2 and L3 fire on
sibling groups, and the remaining graph has no siblings sharing
parents, so both are no-ops. Trajectory shrinks from 6 messages to 4.

\begin{figure}[!htb]
\centering
\begin{tikzpicture}[
  every node/.style={font=\scriptsize},
  rd/.style={circle,draw=black,fill=white,minimum size=4.5mm,inner sep=0pt},
  drop/.style={circle,draw=warn,fill=warn!15,minimum size=4.5mm,inner sep=0pt,text=warn,dashed,line width=0.5pt},
  ->,>={Stealth[length=3.5pt]},
  ba/.style={font=\scriptsize\bfseries,text=gray!40!black}
]
\node[ba,anchor=south] at (0.5, 1.3)  {annotated DAG};
\node[rd] (aR1) at (0.5, 1.0)   {R1};
\node[drop] (aR2) at (1.15, 0.4)  {R2};
\node[rd] (aR3) at (0.5, 0.4)   {R3};
\node[rd] (aR4) at (0.5,-0.2)   {R4};
\draw (aR1) -- (aR2); \draw (aR1) -- (aR3); \draw (aR3) -- (aR4);

\draw[->,thick,gray!60!black,line width=0.7pt] (1.65,0.4) -- (2.20,0.4)
      node[midway,above=-0.5pt,font=\scriptsize\itshape,text=gray!60!black]{L1};

\node[ba,anchor=south] at (2.7, 1.3)  {after L1};
\node[rd] (bR1) at (2.7, 1.0)   {R1};
\node[rd] (bR3) at (2.7, 0.4)   {R3};
\node[rd] (bR4) at (2.7,-0.2)   {R4};
\draw (bR1) -- (bR3); \draw (bR3) -- (bR4);
\end{tikzpicture}
\end{figure}

\paragraph{Sample~C (\texttt{\_idx=13}), L3 Only.}
The annotated DAG has edges \texttt{R1$\to$R2}, \texttt{R1$\to$R4},
\texttt{R2$\to$R3}, \texttt{R3$\to$R5}, \texttt{R4$\to$R5}. No round
is a deletable leaf, so \textbf{L1} is a no-op. For \textbf{L2},
R2's children are $\{\text{R3}\}$ while R4's are $\{\text{R5}\}$,
so the strict criterion (same parents \emph{and} children) does not
fire either. \textbf{L3} requires only shared parents, and R2 and R4
both have parent $\{\text{R1}\}$, so they merge into one round
\texttt{R2+4}. The remaining edges become
\texttt{R1$\to$R2+4}, \texttt{R2+4$\to$R3}, \texttt{R2+4$\to$R5},
\texttt{R3$\to$R5}. Trajectory shrinks from 8 messages to 6.

\begin{figure}[!htb]
\centering
\begin{tikzpicture}[
  every node/.style={font=\scriptsize},
  rd/.style={circle,draw=black,fill=white,minimum size=4.5mm,inner sep=0pt},
  mrg/.style={ellipse,draw=mrgnode,fill=mrgnode!12,minimum width=10mm,minimum height=4.5mm,inner sep=0pt,text=mrgnode!60!black,line width=0.6pt},
  ->,>={Stealth[length=3.5pt]},
  ba/.style={font=\scriptsize\bfseries,text=gray!40!black}
]
\node[ba,anchor=south] at (0.5, 1.5)  {annotated DAG};
\node[rd] (cR1) at (0.5, 1.2)   {R1};
\node[rd] (cR2) at (0.05, 0.55)  {R2};
\node[rd] (cR4) at (0.95, 0.55)  {R4};
\node[rd] (cR3) at (0.05,-0.05) {R3};
\node[rd] (cR5) at (0.5,-0.65)  {R5};
\draw (cR1) -- (cR2); \draw (cR1) -- (cR4);
\draw (cR2) -- (cR3); \draw (cR3) -- (cR5); \draw (cR4) -- (cR5);

\draw[->,thick,gray!60!black,line width=0.7pt] (1.65,0.4) -- (2.20,0.4)
      node[midway,above=-0.5pt,font=\scriptsize\itshape,text=gray!60!black]{L3};

\node[ba,anchor=south] at (2.85, 1.5)  {after L3};
\node[rd] (dR1) at (2.85, 1.2)   {R1};
\node[mrg] (dR24) at (2.85, 0.55) {R2+4};
\node[rd] (dR3) at (2.40,-0.05) {R3};
\node[rd] (dR5) at (2.85,-0.65) {R5};
\draw (dR1) -- (dR24); \draw (dR24) -- (dR3); \draw (dR24) -- (dR5);
\draw (dR3) -- (dR5);
\end{tikzpicture}
\end{figure}

\noindent Together with Sample~B (\texttt{\_idx=1581}) above, the three
samples isolate the three edit primitives. Real trajectories often
chain them, as Sample~B does (L1 then L2a).

\clearpage
\section{Prompt Templates}
\label{app:prompts}
This appendix gathers LLM prompts used in the paper:

\begin{itemize}[leftmargin=12pt,itemsep=1pt,topsep=2pt]
  \item \textbf{Prompt A, DAG annotation}: extracts the round-level
  dependency DAG used by all our edits (Sec.~\ref{sec:dag}).
  \item \textbf{Prompt B, Critical-Path KEEP/REMOVE}: the
  LLM-deletion baseline against which we compare (Sec.~\ref{sec:setup}).
  \item \textbf{Prompt C, Merge-think rephrase}: turns
  raw-concatenated \texttt{<think>} blocks into a single coherent
  passage for \textbf{L2a}/\textbf{L3a} (Sec.~\ref{sec:rephrase}).
  \item \textbf{Prompt D, LLM-as-judge}: scores the model's final
  \texttt{<answer>} against the gold answer. Used as the task-accuracy
  metric in all tables (Sec.~\ref{sec:setup}).
\end{itemize}

\begin{figure}[!htb]
\begin{strictpromptbox}[Prompt A: DAG annotation]
\# Role\\
You are an expert in logical reasoning and causal analysis. Your task
is to analyze a multi-round AI Agent trajectory and deconstruct it
into a Directed Acyclic Graph (DAG) that captures \textbf{only the
dependencies that actually matter for producing the final answer}.

\medskip
\# Task Description\\
The trajectory is organized by ``Rounds.'' You will be told which
round contains the final answer (typically the last round). Your job
is to decide, for every candidate edge \texttt{Round i -> Round j},
whether Round j would have been \emph{impossible or clearly wrong}
without Round i's concrete contribution --- judged \textbf{globally
against the final answer}, not by local narrative flow.

\medskip
\# Definition of a Dependency Edge (Round i -> Round j)\\
An edge exists \textbf{only if ALL of the following are true}:\\
1. \textbf{Concrete carry-over}: A specific fact, number, entity, URL,
identifier, or inferred conclusion that Round j actually \emph{uses}
was first produced in Round i --- and cannot be obtained from any
other earlier round or from the original query alone.\\
2. \textbf{Globally load-bearing}: Removing Round i would break Round
j's ability to make progress toward the final answer in Round N. If
Round j could have been produced by skipping Round i (perhaps with
minor rewording), do NOT add the edge.\\
3. \textbf{Not mere narrative / chronological continuity}: Do NOT add
an edge just because Round j's \texttt{<think>} text mentions, reacts
to, or rhetorically references Round i.

\medskip
\# Anti-Patterns --- Do NOT Add an Edge When:\\
\textbullet\ \textbf{Dead-end rounds}: Round i returned information
not used anywhere downstream (including Round N).\\
\textbullet\ \textbf{Redundant confirmation}: Round j merely
re-verifies a fact already established in an earlier round.\\
\textbullet\ \textbf{Parallel independent lookups}: Round i and j are
both sub-queries derived from a common ancestor, with no information
flowing from one to the other.\\
\textbullet\ \textbf{Pure stylistic/continuity reference}: ``continuing
from the previous step'' without consuming any output of Round i.

\medskip
\# Be Aggressive About Pruning Edges\\
Default to NOT adding an edge. When in doubt, ask: ``If I deleted
Round i from the transcript, would Round j still reach the same
conclusion (perhaps via trivial rewording)?'' If yes --- do not add
the edge.

\medskip
\# Output Format\\
JSON edge list, each entry naming the concrete fact/artifact carried:\\
\texttt{[\{"source": "Round X", "target": "Round Y",}\\
\texttt{\ \ "reason": "Round X produced <fact> which}\\
\texttt{\ \ Round Y uses to <purpose>"\}, ...]}

\medskip
\# Input Trajectory\\
\{INPUT\_TRAJ\_STRING\}
\end{strictpromptbox}
\end{figure}

\begin{figure}[!htb]
\begin{cpathpromptbox}[Prompt B: Critical-Path KEEP/REMOVE baseline (English translation of the production prompt)]
\textbf{System.} You are a rigorous, conservative cleaning assistant
for multi-modal Agent research data. Your task is to prune
unnecessary reasoning / tool-call rounds so the training data is more
efficient, while keeping the remaining trajectory logically coherent
and self-consistent. Output compact JSON only; do not wrap in
markdown code fences.

\medskip
\textbf{User.}\\
\# Background\\
Below is one Agent trajectory. Each middle round consists of an
\textit{assistant} turn (with \texttt{<think>} and \texttt{<tool\_call>})
followed by a \textit{user} turn (\texttt{tool\_response}). The
final assistant turn emits \texttt{<answer>}.

\medskip
\# Goal\\
Decide which middle rounds can be \textbf{deleted entirely}, such
that the trajectory becomes more concise without changing the
correctness of the final answer.

\medskip
\# Typical removable rounds\\
1. \textbf{Failed / empty result}: tool\_response is ``Page not found'',
empty, or irrelevant, and no later round adjusts strategy based on
it.\\
2. \textbf{Redundant repeat}: this tool\_call overlaps a prior one,
information already obtained.\\
3. \textbf{Abandoned branch}: this round tried a direction but no
later think/tool\_call uses its output, and the final answer is
unrelated.\\
4. \textbf{Extra re-verification}: the key fact already entered the
answer via an earlier round; this round only reconfirms.\\
5. \textbf{Over-long reasoning prep}: the tool\_call output of this
round does not appear in the evidence chain for the final answer.

\medskip
\# Non-removable rounds\\
Concrete artefacts of this round (number, year, name, link, image
description, \dots) appear \textbf{directly in the final answer}; OR
some kept round's think / tool\_call explicitly references this
round's output; OR this round is a non-skippable step in the
reasoning chain.

\medskip
\# Coherence hard constraint\\
After deleting a round, the next kept round's \texttt{<think>} prefix
must not read as a reference to the deleted content (e.g., ``let's
try another link'', ``that didn't work'', ``another search'',
``hmm''). To repair such \emph{dangling references}, you may supply a
\textbf{minimal rewrite} (\texttt{patch\_think}) for the affected
kept round. The rewrite must preserve its conclusion and subsequent
tool\_call, only altering the opening one or two sentences.
\texttt{null} means no rewrite needed.

\medskip
\# Conservative principle\\
When in doubt, \textbf{keep} the round. If no round can be safely
deleted, return \texttt{remove: []}.

\medskip
\# Output Format\\
Strict JSON, no markdown:\\
\texttt{\{}\\
\texttt{\ \ "status": "reduce" | "non\_reducible" | "illogical",}\\
\texttt{\ \ "remove": [<round\_id\_int>, ...],}\\
\texttt{\ \ "patch\_think": \{ "<id>": "<rewritten think>", ... \},}\\
\texttt{\ \ "reason": "<short justification>"}\\
\texttt{\}}\\
\textbullet\ \texttt{reduce}: at least one deletable round.\\
\textbullet\ \texttt{non\_reducible}: all rounds load-bearing or
patch\_think cannot fix coherence.\\
\textbullet\ \texttt{illogical}: trajectory itself is broken (answer
mismatched, final answer missing, \dots).\\
\textbullet\ \texttt{remove} must not include Round 1 or the final
answer round.

\medskip
\# Input Trajectory\\
\{INPUT\_TRAJ\_STRING\}
\end{cpathpromptbox}
\end{figure}

\begin{figure}[!htb]
\begin{mergepromptbox}[{Prompt C: Merge-think rephrase (used by \textbf{L2a}, \textbf{L3a})}]
Your task is to merge multiple \texttt{<think></think>} segments from
a trajectory of model outputs into a single coherent reasoning
passage.

\medskip
Context: These \texttt{<think>} segments originally came from
separate turns in a multi-turn user--assistant conversation, where
each \texttt{<think>} block was followed by tool\_call arguments. It
has now been determined that the think contents and the tool\_calls
can each be consolidated into a single turn. Your job is to merge the
think segments into one logical, coherent reasoning passage.

\medskip
Requirements:\\
\textbullet\ Preserve the original intent and logical flow of the
reasoning.\\
\textbullet\ If the original think segments reference or describe
tool\_calls (tool invocations), you must retain that meaning --- do
not drop references to which tools are being used or why.\\
\textbullet\ Smooth out the transitions: the current think contents
are crudely concatenated with semicolons (\texttt{;}) and line
breaks. Rewrite them so they read as one continuous, natural chain of
thought rather than disjoint fragments.\\
\textbullet\ Do not add new reasoning that wasn't present in the
original; only restructure and connect what is already there.\\
\textbullet\ Compress for token efficiency without losing information.
Eliminate redundancy, repeated context, filler phrases, and verbose
restatements. Merge overlapping points --- but every distinct piece
of information, decision, and tool-call reference must still be
present in the output.

\medskip
Input (concatenated think contents):\\
\{CONCATED\_THINK\}

\medskip
Output format: Plain text only. Do not wrap the output in any special
tags (no \texttt{<think>}, no XML, no markdown code fences).
\end{mergepromptbox}
\end{figure}

\begin{figure}[!htb]
\begin{judgepromptbox}[Prompt D: LLM-as-judge (gpt-5-nano)]
Your job is to look at a question, a gold target, and a predicted
answer, and then assign a grade of either \texttt{[CORRECT, INCORRECT,
NOT\_ATTEMPTED]}. First, examples of each grade; then a new example.

\medskip
\textbf{CORRECT examples} (predicted answer is equivalent to the
gold target):\\
Q: ``What are the names of Barack Obama's children?''; Gold: ``Malia
Obama and Sasha Obama''. Accept: ``sasha and malia obama''; ``most
people would say Malia and Sasha, but I'm not sure\dots''; ``\dots
Malia Ann and Natasha Marian, but commonly Malia and Sasha\dots''.
A predicted answer is CORRECT iff it fully contains the gold
information, contradicts nothing in it, and only semantic meaning
matters (case / punctuation / order do not). Hedging is allowed as
long as the gold is fully included.

\medskip
\textbf{INCORRECT examples} (predicted answer contradicts the gold):\\
``Malia.''; ``Malia, Sasha, and Susan.''; ``Obama has no children.'';
``either Malia and Sasha. Or Malia and Jackie\dots''. A factual
statement contradicting the gold is INCORRECT, even if hedged.

\medskip
\textbf{NOT\_ATTEMPTED examples} (predicted answer neither contains
nor contradicts the gold):\\
``I don't know.''; ``I need more context.''; ``He has two children. I
know one is Malia, but I'm not sure about the other.''

\medskip
\textbf{Additional rules:}\\
\textbullet\ Numbers must be correct to the last significant figure
in the gold (``120k''$\to$ accept 115k--124k; reject 100k or 113k).\\
\textbullet\ Gold may carry extra info beyond the question; predicted
need only cover what the question asked.\\
\textbullet\ Do not punish omissions clearly inferred from the
question (``San Francisco'' for ``San Francisco, California'').\\
\textbullet\ Tolerate typos in names if clearly the same person.

\medskip
Here is a new example. Simply reply with \texttt{A}, \texttt{B}, or
\texttt{C} (no other text).\\
\texttt{Question: \{query\}}\\
\texttt{Gold target: \{reference\_answer\}}\\
\texttt{Predicted answer: \{generated\_answer\}}

\medskip
Grade as one of:\\
\texttt{A: CORRECT}\\
\texttt{B: INCORRECT}\\
\texttt{C: NOT\_ATTEMPTED}\\
Return only the single letter.
\end{judgepromptbox}
\end{figure}

\end{document}